\documentclass[10pt,twocolumn,letterpaper]{article}

\usepackage{iccv}
\usepackage{times}
\usepackage{epsfig}
\usepackage{graphicx}
\usepackage{subcaption}
\usepackage{booktabs}
\usepackage[10pt]{moresize}
\usepackage{multirow}
\usepackage{cite}

\usepackage{tikz}
\usepackage{color}
\usepackage{bm}
\usepackage{comment}
\usepackage{amsmath,amssymb} 
\DeclareMathOperator*{\argmax}{argmax}
\usepackage[pagebackref=true,breaklinks=true,letterpaper=true,colorlinks,bookmarks=false]{hyperref}
\usepackage{fancyhdr}
\usepackage[accsupp]{axessibility}  
\iccvfinalcopy 

\usepackage[capitalize]{cleveref}
\crefname{section}{Sec.}{Secs.}
\Crefname{section}{Section}{Sections}
\Crefname{table}{Table}{Tables}
\crefname{table}{Tab.}{Tabs.}

\def\iccvPaperID{5049} 

\begin{document}

\title{
3D-Aware Neural Body Fitting for Occlusion Robust 3D Human Pose Estimation
} 
\author{Yi Zhang\textsuperscript{1}$^*$ \qquad Pengliang Ji\textsuperscript{2}$^*$ \qquad Angtian Wang\textsuperscript{1}\qquad Jieru Mei\textsuperscript{1} \\
 Adam Kortylewski\textsuperscript{3,4} \qquad Alan Yuille\textsuperscript{1} \\
\textsuperscript{1}Johns Hopkins University \qquad \textsuperscript{2}Beihang University \\
\textsuperscript{3}Max Planck Institute for Informatics \qquad \textsuperscript{4}University of Freiburg
}
\maketitle

\def\thefootnote{$*$}\footnotetext{Indicates equal contribution.}

\begin{abstract}
Regression-based methods for 3D human pose estimation directly predict the 3D pose parameters from a 2D image using deep networks. While achieving state-of-the-art performance on standard benchmarks, their performance degrades under occlusion.
In contrast, optimization-based methods fit a parametric body model to 2D features in an iterative manner. The localized reconstruction loss can potentially make them robust to occlusion, but they suffer from the 2D-3D ambiguity. Motivated by the recent success of generative models in rigid object pose estimation, we propose 3D-aware Neural Body Fitting (3DNBF) - an approximate analysis-by-synthesis approach to 3D human pose estimation with SOTA performance and occlusion robustness. In particular, we propose a generative model of deep features based on a volumetric human representation with Gaussian ellipsoidal kernels emitting 3D pose-dependent feature vectors. The neural features are trained with contrastive learning to become 3D-aware and hence to overcome the 2D-3D ambiguity. Experiments show that 3DNBF outperforms other approaches on both occluded and standard benchmarks. Code is available at \url{https://github.com/edz-o/3DNBF}

\end{abstract}

\section{Introduction}

\begin{figure}
    \centering
    \includegraphics[width=\linewidth]{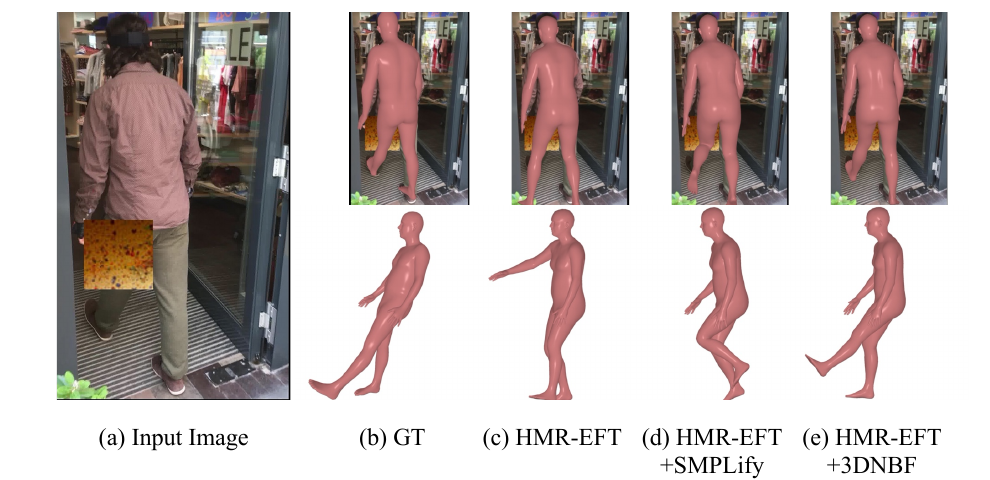}
    \caption{3D human pose estimation under occlusion. Performance of regression-based methods~\cite{joo2021exemplar} degrades under occlusion (c). Traditional optimization-based methods can be robust occlusion, but they suffer from the 2D-3D ambiguity in monocular 3D HPE (d). Our generative approach resolves the 2D-3D ambiguity through analysis-by-synthesis in a 3D-aware feature space (e).
    }
    \label{fig:intro}
\end{figure}

Monocular 3D human pose estimation (HPE) is a long-standing problem of computer vision. 
Regression-based methods \cite{guler_2019_CVPR,kanazawa_hmr,omran2018nbf,pavlakos2018humanshape,Tan,tung2017self} directly regress the 3D pose parameters of a human body model, such as SMPL~\cite{looper_smpl}, and learn to overcome the inherent 2D-3D ambiguity of the prediction task from the training data. 
However, the performance of regression-based methods degrades when humans are partially occluded, as demonstrated by related work \cite{kocabas2021pare} and in our experiments (Figure \ref{fig:intro} (c)). Optimization-based methods~\cite{SMPL-X:2019,SPIN:ICCV:2019,bogo2016keep,xiang2019monocular,Xu_2019_ICCV} fit a parametric body model to 2D representations, such as keypoint detections \cite{kanazawa_hmr,kolotouros2019learning} or segmentation maps~\cite{omran2018nbf,pavlakos2018humanshape,zanfir2020weakly}, in an iterative manner. They are relatively robust to occlusion but perform worse than regression-based methods in 3D HPE particularly because they suffer from the 2D-3D ambiguity (Figure \ref{fig:intro}(d)), even when regularized with strong 3D priors \cite{SMPL-X:2019}, because the manually designed 2D features lack 3D information. 

Recently, generative models have been shown to be successful with improved robustness to occlusion in object recognition~\cite{kortylewski2020compositionalijcv} and rigid object pose estimation~\cite{wang2021nemo} for certain object categories. 
The idea is to formulate vision tasks as inverse graphics or analysis-by-synthesis~\cite{kersten2004object,yuille2006vision} - searching for the parameters in a generative model (e.g. computer graphics models) that best explain the observed image while an outlier process can be introduced to explain occluded regions. However, performing analysis-by-synthesis in RGB pixel space is challenging due to the lack of both good generative models and efficient algorithms to inverse them. Instead, they perform approximate analysis-by-synthesis in deep feature space. However, the generative models used are 2D-based or simple cuboid-like 3D structures with features invariant to the 3D viewpoint, making them less suitable for 3D HPE. 

In this work, we propose a 3D-aware Neural Body Fitting (3DNBF) framework that enables feature-level analysis-by-synthesis for 3D HPE, which is highly robust to occlusion (Figure \ref{fig:intro}(e)). Specifically, we propose a novel generative model of deep network features for human body, named Neural Body Volumes (NBV). NBV is an explicit volume-based parametric body representation consisting of a set of Gaussian ellipsoidal kernels that emit feature vectors. 
Compared with the popular mesh representation, our volume representation is analytically differentiable, provides smooth gradients, i.e. is efficient to optimize, and rigorously handles self-occlusion~\cite{rhodin2015versatile}. We employ a factorized likelihood model for feature maps which is further made robust to partial occlusion by incorporating robust loss functions~\cite{huber2004robust}. To overcome the 2D-3D ambiguity, we impose a distribution on the kernel features conditioned on pose parameters making them pose-dependent.

Unlike optimization-based methods that manually design the feature representation which may lose information, we learn the features from data. 
In particular, we introduce a contrastive learning framework \cite{he2019momentum,wu2018unsupervised,bai2023coke} to learn features that are invariant to instance-specific details (such as color of the clothes), meanwhile encouraging them to capture local 3D pose information of the human body parts, i.e. being 3D-aware. The generative model is learned with the feature extractor network iteratively.
For more efficient inference, we attach a regression head to the feature extractor to predict the pose and shape parameters from the feature directly. 
During inference, we initialize NBV with the prediction from our regressor head and optimize the human pose by maximizing the likelihood of the target feature map under the generative model using gradient-based optimization. 
We find this combined approach can resolve common errors of regression-based methods, such as when the pose of partially occluded parts is not estimated correctly (Figure \ref{fig:intro}).

We evaluate 3DNBF on three existing 3D HPE datasets: \texttt{3DPW}~\cite{von2018recovering}, \texttt{3DPW-Occ}~\cite{zhangoohcvpr20} and \texttt{3DOH50K}~\cite{zhangoohcvpr20}, and propose a more challenging adversarial evaluation protocol \texttt{3DPW-AdvOcc} for occlusion robustness. Our experimental results show that 3DNBF outperforms state-of-the-art (SOTA) regression-based methods as well as optimization-based approaches by a large margin under occlusion while maintaining SOTA performance on non-occluded data.
In summary, our main contributions are:
\begin{enumerate}
\item We propose 3DNBF - an approximate analysis-by-synthesis approach for 3D HPE at feature level with a volume-based neural generative model NBV for human body with pose-dependent kernel features. 
\item We introduce a contrastive learning framework to train NBV with a feature extractor such that the feature activations capture the local 3D-pose information of the body parts, to resolve the 2D-3D ambiguity.
\item We demonstrate on four datasets that 3DNBF outperforms SOTA regression-based and optimization-based methods, particularly when under occlusion.
\end{enumerate}

\section{Related Work}
\noindent\textbf{Monocular 3D Human Pose Estimation.} Existing approaches can be categorized into regression-based and optimization-based methods.
Regression-based methods~\cite{guler_2019_CVPR,kanazawa_hmr,omran2018nbf,pavlakos2018humanshape,Tan,tung2017self} directly estimate 3D human pose from RGB image using a deep network. Different 3D human pose representations are adopted such as 3D joint locations~\cite{mehta2017monocular,rogez2017lcr}, 3D heatmaps~\cite{Pavlakos17,sun2018integral,zhou2017towards} and parameters of a parametric human body~\cite{kanazawa_hmr,pavlakos2018humanshape,kolotouros2019learning}. 
Optimization-based methods~\cite{SMPL-X:2019,SPIN:ICCV:2019,bogo2016keep,xiang2019monocular,Xu_2019_ICCV} involve parametric human models like SMPL~\cite{SMPL-X:2019,scape,looper_smpl}, and produce both the 3D human pose and human shape.
The representative method is SMPLify~\cite{bogo2016keep}, which fits the SMPL model to 2D keypoint detections with strong priors. Exploiting more information into the fitting procedure has been investigated, including silhouettes~\cite{lassner_up3d}, multi-view~\cite{huang_mvsmplify}, more expressive shape models~\cite{totalcapture}. \cite{xiang2019monocular} propose to fit 3D part affinity maps to overcome 2D-3D ambiguity. This requires the network to learn accurate part orientation which is difficult and shown to be less robust to occlusion. Hybrid methods~\cite{corona2022learned, song2020human} perform iterative optimization using regressed descent directions. 

\noindent\textbf{Robustness to Occlusion.} Regression-based methods are sensitive to occlusions as studied by Kocabas \etal~\cite{kocabas2021pare}, who propose part segmentation guided attention mechanism to handle occlusion. 
Data augmentation is another common way to enhance occlusion robustness, for example by cropping~\cite{biggs2020multibodies,joo2021exemplar,Rockwell2020}, or by putting patches into the image~\cite{georgakis2020hierarchical,sarandi2018robust}. Even with data augmentation, we show that they are still sensitive to occlusion by applying a more sophisticated sliding-window attack. Explicit occlusion handling in regression-based methods infers occluded joints using representation redundancy~\cite{mehta2018single,mehta2020xnect} which is partially successful or exploits visibility information in the training~\cite{cheng2019occlusion,wang20203d,zhangoohcvpr20,liu2022explicit}. However, the occlusion information except self-occlusion is often unavailable in the wild and expensive to annotate which limits the applicability of such methods. To model pose ambiguities for truncated human images, \cite{biggs2020multibodies,kolotouros2021probabilistic,sengupta2021hierarchical} predict multiple possible poses that have correct 2D projections. 
Another direction to handle occlusion leverages motion in sequences~\cite{cheng20203d,huang2022occluded}. 
Generative models are shown to be robust to occlusion~\cite{moreno2016overcoming,wang2021nemo} for parsing rigid objects and we further demonstrate it for articulated objects. 

\noindent\textbf{Human body representations.} Parametric mesh models~\cite{SMPL-X:2019,SPIN:ICCV:2019,bogo2016keep,xiang2019monocular,Xu_2019_ICCV} are most popular models for human pose/shape estimation, which generate intermediate representations like 2D keypoints, sihouette and part segmentations. However, these representations lose the local information, e.g. shading, useful for inferring 3D from 2D. Recently, implicit volume representation has become increasingly popular~\cite{saito2019pifu,liu2021neural,peng2021neural,su2021nerf,park2021nerfies,noguchi2021neural,xiu2022icon,jiang2022neuman,weng2022humannerf} as they can achieve highly realistic human reconstruction. However, they are not suitable for our purpose as training these models often requires multi-view or videos and takes an extended time for a single person. We propose a body representation that combines a volumetric 3D Gaussian representation~\cite{robertini2014efficient,rhodin2015versatile} which gives more stable gradients compared to mesh-based differentiable rendering. Compared to popular implicit volume representations, our volume representation is explicit with fewer parameters to learn which leads to efficient inference.

\noindent\textbf{Generative Models of Neural Textures.} Prior works have shown the potential of combining 3D representations with neural texture maps, with an application to image synthesis of static scenes through neural rendering \cite{nguyen2019hologan,thies2019deferred,Niemeyer2020GIRAFFE}. As inverting a generative model of RGB pixel values is challenging, a recent line of work introduced a neural analysis-by-synthesis approach to perform visual recognition tasks such as image classification \cite{kortylewski2020compositional,kortylewski2020compositionalijcv,kortylewski2020combining} and 3D pose estimation \cite{wang2021nemo} with a largely enhanced robustness to partial occlusion when compared to standard deep network based approaches. However, these prior works explicitly assume rigid objects and use simple 2D-based or cuboid-like generative models. \cite{neverova2020continuous} learns a continuous feature embedding function for each vertex on 3D human body mesh. However, this representation is invariant to 3D pose and therefore loses the useful information for estimation 3D from 2D. 
Our work generalizes the neural analysis-by-synthesis approach to 3D HPE, addressing the challenge of 2D-3D ambiguities and modeling articulated human bodies.
\section{3D-Aware Neural Body Fitting}
In the following, we first explain a conceptual formulation of analysis-by-synthesis for 3D human pose estimation (Section \ref{sec:abs}) and propose our feature-level analysis-by-synthesis formulation in Section~\ref{sec:fabs}. Then we introduce the proposed generative Neural Body Volume model (Section \ref{sec:nbv}), including the 3D-aware pose-dependent features (Section \ref{method:3d-aware}). Finally, we describe the training and inference process for the generative model in Section \ref{sec:train}, \ref{sec:inference}.

\subsection{HPE via Analysis-by-Synthesis}\label{sec:abs}
Given an input image $\bm{I} \in \mathbb{R}^{H\times W \times 3}$, we aim to estimate the 3D human pose parameters $\bm{\theta}$.
Using Bayes rule we formulate the pose estimation task as a probabilistic inference problem given the observed image $\bm{I}$:
\begin{equation}
    \bm{\theta}^* = \argmax_{\bm{\theta}} p(\bm{\theta}|\bm{I}) 
    =\argmax_{\bm{\theta}} p(\bm{I}|\bm{\theta})p(\bm{\theta}), \label{eq:abs}
\end{equation}
where $p(\bm{\theta})$ is a prior distribution learned from data~\cite{bogo2016keep,SMPL-X:2019}, and $p(\bm{I}|\bm{\theta})$ is the likelihood. 
$p(\bm{I}|\bm{\theta})$ is typically defined using a generative forward model (involving 3D CAD models and a graphics engine)
, and the analysis-by-synthesis process is hence defined as finding the parameters $\bm{\theta}^*$ that can best explain the input image. 
However, it is very challenging to reconstruct human images accurately which requires either multi-view images or video input~\cite{liu2021neural,peng2021neural,su2021nerf,jiang2022neuman,weng2022humannerf}. 

\begin{figure*}[t]
    \centering
    \includegraphics[width=1\linewidth]{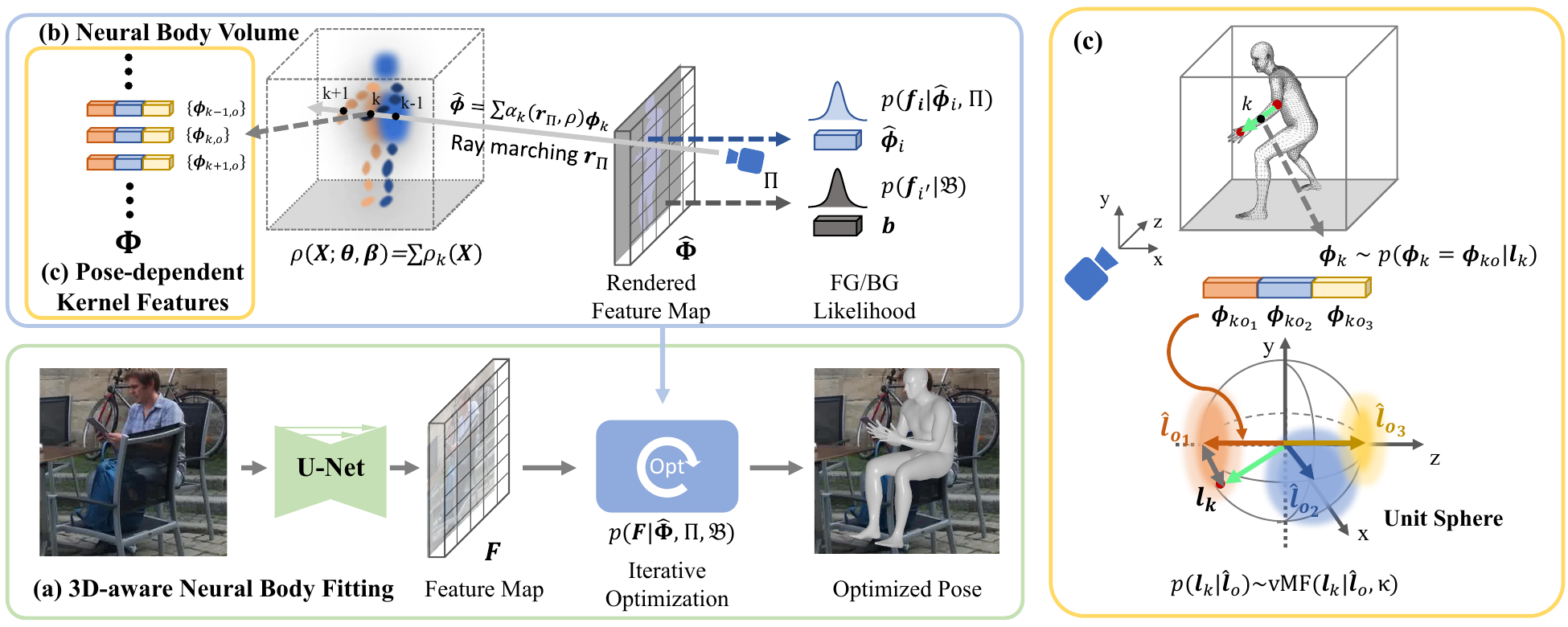}
    \caption{\textbf{Overview of our system.} (a) We perform feature-level analysis-by-synthesis for 3D human pose estimation by fitting a 3D-aware generative model of deep feature (NBV) to the feature map $\bm{F}$ extracted by a U-Net. (b) NBV is defined as a volume representation of human body $\rho$, driven by pose and shape parameters $\{\bm{\theta}, \bm{\beta}\}$, which consists of a set of Gaussian kernels each emitting a pose-dependent feature $\bm{\phi}$. Volume rendering is used to render NBV to a feature map $\bm{\hat{\Phi}}$. The foreground feature likelihood is defined as a Gaussian distribution centered at the rendered feature vector while the background feature likelihood is modeled by a background model. Pose estimation is done by optimizing the negative log-likelihood (NLL) loss of $\bm{F}$ w.r.t. $\{\bm{\theta}, \bm{\beta}\}$ and camera $\Pi$. (c) the distribution of the kernel feature is conditioned on the orientation of the limb that the kernel belongs to. }
    \label{fig:system_overview}
\end{figure*}

Instead of performing analysis-by-synthesis in RGB space, we aim to reconstruct the human appearance at the \textit{feature-level} of a neural network.
Fig.~\ref{fig:system_overview} is an overview of our method. The feature representations will be learned to become invariant to image variations that is not relevant for the HPE task, such as clothing color or style, and hence will enable us to perform HPE accurately from a single image. 
In the following, we will first introduce the concept of feature-level analysis-by-synthesis and subsequently introduce a generative model of humans on the feature level.

\subsection{Feature-Level Analysis-by-Synthesis}\label{sec:fabs}
We denote a feature representation of an input image as $\zeta(\bm{I}) = \bm{F} \in \mathbb{R}^{H\times W \times D}$ which is the output of a deep convolutional neural network $\zeta$. 
$\bm{f}_i \in \mathbb{R}^D$ is a feature vector in $\bm{F}$ at pixel $i$ on the feature map. 
We define a generative model of humans on the feature-level as $\mathcal{G}(\bm{\theta})=\bm{\hat{\Phi}} \in \mathbb{R}^{H\times W \times D}$, which produces a feature map $\bm{\hat{\Phi}}$ given the pose $\theta$. 
We can now define the likelihood function of our Bayesian model (Eq.~\ref{eq:abs}). 
To enable efficient learning and inference, we adopt a factorized likelihood model:
{\small
\begin{equation}
    p(\bm{F}|\mathcal{G}(\bm{\theta}), \mathcal{B})=\prod_{i\in\mathcal{FG}}p(\bm{f_i}|\bm{\hat{\phi}_i})\prod_{i'\in\mathcal{BG}}p(\bm{f_{i'}}|\mathcal{B}), 
    \label{eq:prob1}
\end{equation}
}where the foreground $\mathcal{FG}$ is the set of all pixel locations on the feature map $\bm{F}$ that are covered by the human. 
The background $\mathcal{BG}$ contains those pixels respectively that are not covered. 
The foreground likelihood $p(\bm{f_i}|\bm{\hat{\phi}_i})$ is defined as a Gaussian distribution $\mathcal{N}(\bm{\hat{\phi}_i}, \sigma_i^2\bm{I})$ with the mean vector $\bm{\hat{\phi}_i}$ at location $i$, and a standard deviation $\sigma_i$. 
Background features are modeled using a simple background model $p(\bm{f_{i'}}|\mathcal{B})$ that is defined by a Gaussian distribution $\mathcal{N}(\bm{b}, \sigma^2\bm{I})$, where the parameters are $\mathcal{B}=\{\bm{b}, \sigma\}$ learned from the background features in the training images.

\noindent\textbf{Occlusion robustness.} Following related work on occlusion-robust analysis-by-synthesis \cite{egger2018occlusion}, we define a robust likelihood as:
{\small
\begin{align}
\label{eq:rec}
    &p(\bm{F}|\mathcal{G}(\bm{\theta}), \mathcal{B}, \bm{Z}) = \prod_{i \in \mathcal{FG}} p(\bm{f_i}|\bm{\hat{\phi}_i}, z_i)
    \prod_{i'\in \mathcal{BG}} p(\bm{f_{i'}}|\mathcal{B}) \\
    & p(\bm{f_i}|\bm{\hat{\phi}_i}, z_i) = \left[p(\bm{f_i}|\bm{\hat{\phi}_i})\right]^{z_i} \left[p(\bm{f_i}|\mathcal{B})\right]^{(1-z_i)}, \nonumber
\end{align}
}where $z_i\in\{0,1\}$ is a binary variable and we set its prior probabilities to be $p(z_i\texttt{=}1)=p(z_i\texttt{=}0)=0.5$. 
The variable $z_i$ allows the background model to explain those pixels in $\mathcal{FG}$ that cannot be explained well by the foreground model, presumably due to partial occlusion. To reduce clutter in the remaining paper we will omit the occlusion variable in the coming equations, but note that we are using a robust likelihood during inference.

In the following section, we describe our feature-level generative model for human pose estimation. 

\subsection{Neural Body Volumes}
\label{sec:nbv}
At the core of our framework is the \textit{Neural Body Volumes} (NBV) representation, a model that enables the rendering of human bodies on the feature-level (illustrated in Figure~\ref{fig:system_overview}(b)).
Traditional human body models are mostly mesh-based, e.g. SMPL~\cite{looper_smpl}. 
However, while meshes are useful representations for forward-rendering applications in computer graphics, they are sub-optimal for differentiable inverse rendering, since the mesh rendering process is inherently difficult to differentiate w.r.t. the model parameters \cite{liu2019soft}.
Prior art~\cite{rhodin2015versatile} showed that volume rendering has a smoother and analytical gradient, and 
leads to a more efficient optimization, and better handling self-occlusion compared to meshes. 
Inspired by these results, we propose Neural Body Volumes (NBV), a volume-based representation of human bodies for rendering human bodies on the feature-level.
In NBV, a human body is represented by a three-dimensional volume that consists of $K$ Gaussian kernels placed on the body surface. The density at spatial location $\bm{X}\in\mathbb{R}^3$ is $\rho_k(\bm{X}) = \mathcal{N}(\bm{M}_k, \bm{\Sigma}_{k})$.
$\bm{M_k}(\bm{\theta}, \bm{\beta}) \in \mathbb{R}^{3}$ and $\bm{\Sigma}_k(\bm{\theta}, \bm{\beta}) \in \mathbb{R}^{3\times 3}$ are the mean vector and covariance matrix conditioned on human pose  and shape, parameterized by $\bm{\theta}$ and $\bm{\beta}$ respectively, controlling the center and shape of the Gaussian kernel which we describe in detail in the following paragraph. 
The volume density is defined as $\rho (\bm{X}) = \sum_{k=1}^K \bm{\rho}_k(\bm{X})$.
Each Gaussian kernel is associated with a feature vector $\bm{\phi_k} \in \mathbb{R}^{D}$ which can be rendered to image space using volume rendering:
{\small  
\begin{align}
    \bm{\hat{\phi}} (\bm{r}_{\Pi}) = \int_{t_n}^{t_f} T(t) \sum_{k=1}^K \rho_k(\bm{r}_{\Pi}(t))\bm{\phi_k} \mathrm{d}t,\\
    \text{where}\ T(t)=\exp\left(-\int_{t_n}^{t} \rho(\bm{r}_{\Pi}(s))\mathrm{d}s\right), \nonumber
    \label{pro1}
\end{align}
}
that computes the aggregated feature along the ray $\bm{r}_{\Pi}(t)$  $t\in[t_n, t_f]$ from the camera center through a pixel on the image plane where $\Pi$ denotes the camera parameters. The Gaussian kernel representation enables calculation of the analytic form of the integral $\bm{\hat{\phi}}(\bm{r})=\sum_{k=1}^K\alpha_k(\bm{r}_{\Pi}, \rho)\bm{\phi}_k$ which we provide in the supplementary material.
Here, the number of Gaussian kernels $K$ and the associated features $\Phi=\{\phi_k\}$ are global parameters shared across all human instances. While for each input image, we optimize the pose $\bm{\theta}$ and shape $\bm{\beta}$ to transform the location $\bm{M_k}(\bm{\theta}, \bm{\beta})$ and shape $\bm{\Sigma}_k(\bm{\theta}, \bm{\beta})$ of the Gaussian ellipsoids. Our model can fit arbitrary shapes with a sufficient number of kernels.

\noindent\textbf{Conditioning on pose and shape. } Given a set of body joints $\bm{J} \in \mathbb{R}^{N\times 3}$, the pose is defined as their corresponding rotation matrices $\bm{\Omega} \in \mathbb{R}^{N\times 3\times 3}$ relative to the template joints $\bm{\bar{J}}$ in a skeleton tree. We model body articulation using linear blend skinning (LBS)~\cite{lewis2000pose} which transforms the center of the Gaussian kernels with transformation linearly blending the accumulated rigid transformations $\bm{G}(\bm{J},\bm{\Omega})\in \mathbb{R}^{N\times 4 \times 4}$ of the $N$ body joints (including the root transformation). And we model body shape variations by displacing the kernels with linear combinations of a set of $L$ basis shape displacements $\bm{S} \in \mathbb{R}^{L\times K\times 3}$: 
{\small 
\begin{equation}
    \bm{M_k} = \sum_{i=1}^N w_{k,i}\bm{G}_i [\bm{\bar{M}_k}+\sum_{l=1}^L \beta_l\bm{S}_{lk} | \bm{1}],
\end{equation}
}where $\bm{\bar{M}_k}$ denotes the kernel position in rest pose, and $\sum_{i=1}^N w_{k,i}=1$ and $\sum_{l=1}^L\beta_l=1$ are pose and shape blend weights. $[\cdot|\bm{1}]$ denotes the homogeneous coordinates. For the spatial covariance, we also perform transformation and blending according to the rotation of the joints:
{\small
\begin{equation}
    \bm{\Sigma}_k^{-1} = \sum_{i=1}^N w_{k,i}\bm{R}_i^T \bm{\bar{\Sigma}}_k^{-1} \bm{R}_i, \text{ } \bm{R_i}=\prod_{j\in A(i)} \bm{\Omega_i},
\end{equation}
}where $\bm{\bar{\Sigma}_k}$ is the covariance matrix in rest pose, and $A(i)$ is the ordered set ancestors of joint $i$. This takes into account that the orientation of the Gaussian ellipsoid should rotate with the pose.

The template joints location $\bm{\bar{J}}$ can also deform according to shape. Specifically, we regress the template joint locations from the locations of the deformed Gaussian kernels $\bm{\bar{J}}=g(\bm{\bar{M}}+\sum_{l=1}^L \beta_l\bm{S}_l$). The common choice for such regressor $g:\mathbb{R}^{K\times 3}\to\mathbb{R}^{N\times 3}$ is a linear function~\cite{looper_smpl,SMPL-X:2019}. 

In summary, the proposed Neural Body Volume representation enables us to render human bodies on the feature-level using volume rendering process such that for each pixel in the feature map there will be a feature vector $\bm{\hat{\phi}}$ corresponding to the contribution from all Gaussian kernels.

\subsection{A Generative Model of 3D-Aware Features}
\label{method:3d-aware}

Related work on feature-level inverse rendering for rigid pose estimation \cite{wang2021nemo,iwase2021repose} trains the feature extractor $\zeta$ such that the features become invariant to changes in the 3D pose. 
However, for human pose estimation, it is fundamentally important for the feature representation to be 3D-aware, in order to resolve the inherent 2D-3D ambiguity (as shown in Fig.~\ref{fig:intro}).
To resolve this problem, we aim to learn \textit{pose-dependent} feature representations that is able to better resolve the 2D-3D ambiguity of human poses.\\

\noindent\textbf{3D pose-dependent features for NBV. } To overcome the 2D-3D ambiguity, we make the generative model 3D-aware. In particular, we impose a distribution on the kernel features $\bm{\Phi}$ conditioned on the human pose and shape as shown in Fig.~\ref{fig:system_overview}(c). Therefore, the rendered kernel features explicitly carry 3D pose information. 
Specifically, we define a set of body limbs $\{(\bm{J}_i, \bm{J}_j)|(i,j)\in\mathcal{L}\}$ each defined as an ordered tuple connecting two body joints. The orientation of a limb is defined as $\bm{l}=(\bm{J}_j-\bm{J}_i)/\|\bm{J}_j-\bm{J}_i\|$. We first learn to assign each Gaussian kernel in NBV to one limb according to the pose blend weights. Then we associate each kernel with multiple features $\{\bm{\phi}_o\}$ that correspond to a set of predefined limb orientations $\{\bm{\hat{l}_o}\in\mathbb{R}^{3}\}_{o=1}^O, \|\bm{\hat{l}_o}\|=1$. The distribution for the feature vector of the Gaussian kernel $k$ is then defined as: 
{\small
\begin{align}
    &p(\bm{\phi}_k=\bm{\phi}_{ko}|\bm{l}_k(\bm{\theta}, \bm{\beta})) = \frac {p(\bm{l}_k|\bm{\hat{l}}_o)}{\sum_{o=1}^O p(\bm{l}_k|\bm{\hat{l}}_o)}, 
\end{align}
}where $\bm{l}_k(\bm{\theta}, \bm{\beta})\in\mathbb{R}^3$ is the orientation of the limb that Gaussian kernel $k$ belongs to. $p(\bm{l}_k|\bm{\hat{l}}_o)$ is the von Mises-Fisher distribution $\mathrm{vMF}(\bm{l}_k|\bm{\hat{l}}_o, \kappa_o)$. 
In the simple case of only one kernel $k$, the likelihood of feature at foreground pixel $i$ becomes a Gaussian Mixture Model (GMM):
{\small
\begin{align}
p(\bm{f_i}|\bm{\hat{\phi}_i})&=\sum_{o=1}^{O}p(\bm{\phi}_k=\bm{\phi}_{ko}|\bm{l}_k)\mathcal{N}(\bm{\hat{\phi}_{ko}}, \sigma_{io}^2\bm{I}), 
\end{align}
}where $\bm{\hat{\phi}_{ko}}$ is the feature rendered from $\bm{\phi}_{ko}$. 
Intuitively, the rendered feature has different distributions under different 3D limb orientations. Therefore, we can unambiguously infer the 3D pose from the observed features. During inference, we use the expectation of the kernel feature $\mathbb{E}(\bm{\phi}_k|\bm{\theta}, \bm{\beta})$ for volume rendering for differentiability.

\subsection{Training}\label{sec:train}
Given a set of images $\{\bm{I}_n\}_{n=1}^N$, with ground truth 3D keypoints $\{\bm{\hat{J}}_n\}_{n=1}^N$ and shape $\{\mathcal{V}_n\}_{n=1}^N$, we need to learn a set of parameters in NBV: the template Gaussian kernels and the associated features $\{\bm{\bar{M}},\bm{\bar{\Sigma}}, \bm{\Phi}\}$, the template joints $\bm{\bar{J}}$, the blend weights $\bm{W}$, the basis shape displacements $\bm{S}$ and the joint regressor $g$. We also need to train the UNet feature extractor $\zeta$. We train our model in separate steps by first learning the pose/shape-related parameters $\{\bm{\bar{M}},\bm{\bar{\Sigma}}, \bm{\bar{J}}, \bm{W}, \bm{S}, g\}$ then the kernel features $\bm{\Phi}$ and $\zeta$. 

\noindent\textbf{Learning pose and shape parameters in NBV. } Starting from a downsampled version of a template body mesh model created by artists, we initialize the kernel centers $\bm{\bar{M}}$ with the locations of the vertices and compute the spatial covariance matrices $\bm{\bar{\Sigma}}$ based on the distance of the vertices to their neighbors with the desired amount of overlap. Following \cite{looper_smpl}, a manual segmentation of the template mesh is leveraged to obtain the initial template joints $\bm{\bar{J}}$, the linear joint regressor $g$, and the blend weights $\bm{W}$. Then we train all pose-related parameters $\{\bm{\bar{M}}, \bm{\bar{J}}, \bm{W}, g\}$ together with instance-specific pose $\bm{\theta}$ by minimizing the reconstruction error between the Gaussian kernels and the ground truth shape $\mathcal{V}$. After that, the ground truth shapes are transformed back to the rest pose and the shape basis $\bm{S}$ is obtained by running PCA on these pose-normalized shapes. We refer the readers to ~\cite{looper_smpl} for details as we share a similar training process for this part. Another regressor $\hat{g}$ is trained to regress the ground truth keypoints from kernel centers $\bm{M}$. In practice, we can directly convert a trained SMPL model~\cite{looper_smpl} to NBV by placing the Gaussian kernels at the vertices on the SMPL mesh.

After training the pose/shape-related parameters, we register our NBV to the train set to obtain the ground truth shape and pose for each training sample $\{\bm{\theta}_n\}$, $\{\bm{\beta}_n\}$. Then, we learn the NBV kernel features and a UNet feature extractor $\zeta$ jointly in an iterative manner. 

\noindent\textbf{MLE learning of NBV kernel features.} If $\zeta$ is trained, we can learn the kernel features $\bm{\Phi}$ through maximum likelihood estimation (MLE) by minimizing the following negative log-likelihood of the feature representations over the whole training set,
{\small
\begin{align}
    \mathcal{L}_{\text{NLL}}(\bm{\hat{F}}, \bm{\Phi})=-\sum_{i\in\mathcal{FG}} \log p(\bm{\hat{f}}_{i}|\bm{\hat{\phi}}_{i}),
    \label{equ:nbv_nll}
\end{align}} where for training efficiency, we use an approximate solution to avoid matrix inversion $\bm{\phi}_{ko} = \frac{\sum_{i\in\mathcal{K}} \gamma_{iko}\bm{\hat{f}}_{i}}{\sum_{i\in\mathcal{K}} \gamma_{iko}}$, 
where $\mathcal{K}$ is the set of pixels in the training data that the kernel feature $\bm{\phi}_{ko}$ contributes to, and $\gamma_{iko}$ is the contribution of $\bm{\phi}_{ko}$ to pixel $i$ which is obtained from the volume rendering process. Similarly, the parameters of the background distribution are learned using MLE on the features that are not covered by the projected NBV model in the training data. To reduce the computational cost, we follow \cite{wang2021nemo}~and employ a momentum strategy~\cite{he2020momentum} to update $\bm{\Phi}$ in a moving average manner.

\noindent\textbf{3D-aware contrastive learning of the UNet feature extractor.} Given the generative model, we can train the UNet feature extractor with the NLL loss as defined in Equation~\ref{equ:nbv_nll} w.r.t. the network parameters. In addition, we want the extracted feature map to have the property that the rendered feature from NBV in the ground truth pose has the largest probability. To this end, we incorporate a set of contrastive losses:
{\small
\begin{align}
   & \mathcal{L}_{\text{FG}}(\bm{F},\mathcal{FG}) = - \sum_{i \in \mathcal{FG}}\hspace{.1cm}\sum_{i'\in\mathcal{FG}  \setminus \{i\}} \lVert \bm{f}_{i} - \bm{f}_{i'}\rVert^2\\
   & \mathcal{L}_{\text{3D}}(\bm{F}) = - \sum_k \sum_o \sum_{o'\in \mathcal{O} \setminus\{o\}} \sum_{i\in\mathcal{K}_{ko}}\sum_{j\in\mathcal{K}_{ko'}} \lVert \bm{f}_{iko} - \bm{f}_{jko'}\rVert^2\\
   & \mathcal{L}_{\text{BG}}(\bm{F},\mathcal{FG},\mathcal{BG}) = - \sum_{i \in \mathcal{FG}}
    \sum_{j \in \mathcal{BG}} \lVert \bm{f}_{i} - \bm{f}_{j}\rVert^2
\end{align}
}where $\mathcal{L}_{\text{FG}}$ encourages features of different pixels to be distinct from each other. $\mathcal{L}_{\text{3D}}$ encourages features of the same kernel in different 3D poses to be distinct from each other, i.e. to become 3D-aware. $\mathcal{L}_{\text{BG}}$ encourages features on the human to be distinct from those in the background. We optimize those losses jointly $\mathcal{L}_{\text{contrast}} = \mathcal{L}_{\text{FG}}+\mathcal{L}_{\text{3D}}+\mathcal{L}_{\text{BG}}$ in a contrastive learning framework. Therefore, the total loss for training $\zeta$ is $\mathcal{L}_{train}=\mathcal{L}_{\text{NLL}}+\mathcal{L}_{\text{contrast}}$.

\noindent\textbf{Bottom-up initialization with regression heads. } For efficient inference, it is a common practice in generative modeling to initialize with regression-based methods. In our model, we add a regression head to the UNet feature extractor to predict the pose and shape parameters $\{\bm{\theta}, \bm{\beta}\}$ from the observed feature map. The regression head and the UNet are learned jointly. 

\subsection{Inference}\label{sec:inference}
We estimate the 3D human pose $\bm{\theta}$, shape parameters $\bm{\beta}$, and the camera parameters $\Pi$ using the analysis-by-synthesis formulation in Equation~\ref{eq:abs}. This boils down to minimizing an NLL loss plus a regularization term from the pose prior $p(\bm{\theta})$ w.r.t $\{\bm{\theta}, \bm{\beta}, \Pi\}$. The initialization comes from the regression head. Our proposed generative model is fully differentiable and therefore can be optimized using gradient-based methods.

\subsection{Implementation Details}
We convert the neutral SMPL model to NBV using the method described in Sec.~\ref{sec:train}, keeping $858$ kernels. We use a U-Net~\cite{ronneberger2015u} style network as the feature extractor which consists of a ResNet-50~\cite{he2016deep} backbone and 3 upsampling blocks. The regression head follows the design of \cite{kocabas2021pare}. The input image is a $320\times 320$ crop centered around the human. The feature map has a 4$\times$ downsampled resolution and the feature dimension is $64$ which balances performance and computation cost as shown in ablation in Sec.~\ref{subsec:ablation}. The Adam optimizer with a learning rate of $5\times 10^{-5}$ and batch size of $64$ is used for training the feature extractor and the regression head. Standard data augmentation techniques are used including random flipping, scaling, and rotation. 
For the 3D pose-dependent features, we consider the limb orientation projected to the yz-plane and split the unit circle evenly. We set $O\texttt{=}4$ for all kernels which already gives good enough results as shown in the ablation study in Sec.~\ref{subsec:ablation}. We consider 9 limbs including the left/right upper/lower arm/leg and the torso. The torso includes the head and its orientation is defined as the direction from the mid-hip joint to the neck joint. 
For inference, we also use Adam as the optimizer with a learning rate of $0.02$ and run a maximum of $80$ steps. We use VPoser~\cite{SMPL-X:2019} as our 3D pose prior. We check the negative log-likelihood $\mathcal{L}_{NLL}$ of the initial pose and its $180^{\circ}$-rotated version around y-axis and use the better one to initialize our model. Inference speed is $\sim 1.7$fps with a batch size of 32 on 4 NVIDIA Titan Xp GPUs.

\section{Experiments}
\label{sec:exp}
In this section, we demonstrate the effectiveness and robustness of 3DNBF by comparing it with SOTA HPE. In addition to existing benchmarks, we propose a more challenging adversarial evaluation for occlusion robustness. Finally, we conduct ablation studies to verify the design choices and effectiveness of different components.

\begin{table*}[t]
\small
\centering
\setlength\tabcolsep{2pt}
\begin{tabular}{@{}l@{}c@{}c@{}c@{}|c@{}c@{}c@{}|c@{}c@{}c@{}|c@{}c@{}c@{}|c@{}c@{}c@{}}
\toprule
\multirow{2}{*}{Method} & \multicolumn{3}{c|}{\texttt{3DPW}~\cite{von2018recovering}} & \multicolumn{3}{c|}{\texttt{3DPW-Occ}~\cite{zhangoohcvpr20}} & \multicolumn{3}{c|}{\texttt{3DPW-AdvOcc@40}} & \multicolumn{3}{c|}{\texttt{3DPW-AdvOcc@80}} & \multicolumn{3}{c}{\texttt{3DOH50K}~\cite{zhangoohcvpr20}} \\ \cmidrule{2-4}  \cmidrule{5-7}  \cmidrule{8-10} \cmidrule{11-13} \cmidrule{14-16}
    & \multicolumn{1}{c}{\scriptsize MPJPE$\downarrow$} & \multicolumn{1}{c}{\scriptsize P-MPJPE$\downarrow$} & \scriptsize PCKh$\uparrow$ & \multicolumn{1}{c}{\scriptsize MPJPE$\downarrow$} & \multicolumn{1}{c}{\scriptsize P-MPJPE$\downarrow$} & \scriptsize PCKh$\uparrow$ & \multicolumn{1}{c}{\scriptsize MPJPE$\downarrow$} & \multicolumn{1}{c}{\scriptsize P-MPJPE$\downarrow$} & \scriptsize PCKh$\uparrow$ & \multicolumn{1}{c}{\scriptsize MPJPE$\downarrow$} & \multicolumn{1}{c}{\scriptsize P-MPJPE$\downarrow$} & \scriptsize PCKh$\uparrow$ & \multicolumn{1}{c}{\scriptsize MPJPE$\downarrow$} & \multicolumn{1}{c}{\scriptsize P-MPJPE$\downarrow$} & \scriptsize PCKh$\uparrow$ \\ \midrule
SPIN~\cite{kolotouros2019learning} & 96.6 & 58.3 & 91.7 & 97.5 & 60.8 & 85.9 & 203.5 & 97.0 & 63.6 & 338.8 & 111.7 & 34.1 & 101.3 & 67.9 & 83.3 \\ 
HMR-EFT~\cite{joo2021exemplar} & 89.5 & 53.4 & 93.1 & 95.8 & 57.1 & 87.2 & 146.7 & 73.2 & 77.8 & 202.8 & 83.7 & 63.7 & 97.4 & 65.8 & 84.4  \\ 
MGraphr~\cite{lin2021mesh} & 80.4 & 53.4 & 88.7 & 116.8 & 75.7 & 66.6 & 158.8 & 93.2 & 70.8 & 261.5 & 121.0 & 48.8 & 127.4 & 76.0 & 79.8  \\
PARE~\cite{kocabas2021pare} & 81.4 & 50.9 & 92.5 & 86.8 & 58.8 & 86.2 & {126.5} & {72.5} & 82.3 & {210.9} & {97.4} & 61.9 & 100.7 & 65.1 & 84.2  \\ 
3DNBF & \textbf{79.8} & \textbf{49.3} & \textbf{95.7} & \textbf{77.2} & \textbf{51.2} & \textbf{93.1} & \textbf{105.1} & \textbf{60.5} &\textbf{92.0} & \textbf{140.7} & \textbf{71.8} & \textbf{85.0} & \textbf{86.7} & \textbf{57.5} & \textbf{88.6}  \\ 
\bottomrule
\end{tabular}
\caption{\label{tab:regressor}\textbf{Evaluation on \texttt{3DPW}, \texttt{3DPW-Occ}, \texttt{3DPW-AdvOcc}, and \texttt{3DOH50K}.} The number $40$ and $80$ after \texttt{3DPW-AdvOcc} denote the occluder size. Note the performance improvement of 3DNBF increases as occlusion becomes more severe. (P-MPJPE: PA-MPJPE; MGraphr: Mesh Graphormer.)
}
\end{table*}

\begin{table*}[]
\small
\centering
\setlength\tabcolsep{2pt}
\begin{tabular}{@{}l@{}c@{}c@{}c@{}|c@{}c@{}c@{}|c@{}c@{}c@{}}
\toprule
\multirow{2}{*}{Method} & \multicolumn{3}{c|}{\texttt{3DPW}~\cite{von2018recovering}} & \multicolumn{3}{c|}{\texttt{3DPW-AdvOcc@40}} & \multicolumn{3}{c}{\texttt{3DPW-AdvOcc@80}} \\ \cmidrule{2-4}  \cmidrule{5-7}  \cmidrule{8-10}
 & \multicolumn{1}{c}{\scriptsize MPJPE$\downarrow$} & \multicolumn{1}{c}{\scriptsize PA-MPJPE$\downarrow$} & \multicolumn{1}{c}{\scriptsize PCKh$\uparrow$} & \multicolumn{1}{c}{\scriptsize MPJPE$\downarrow$} & \multicolumn{1}{c}{\scriptsize PA-MPJPE$\downarrow$} & \multicolumn{1}{c}{\scriptsize PCKh$\uparrow$} & \multicolumn{1}{c}{\scriptsize MPJPE$\downarrow$} & \multicolumn{1}{c}{\scriptsize PA-MPJPE$\downarrow$} & \multicolumn{1}{c}{\scriptsize PCKh$\uparrow$} \\ \midrule
HMR-EFT~\cite{joo2021exemplar} & 89.5 & 53.4 & 93.1 & 146.7 & 73.2 & 77.8 & 202.8 & 83.7 & 63.7 \\ \midrule
+ SMPLify~\cite{bogo2016keep} & 106.2 & 64.8 & 91.2 & 133.4 & 75.8 & 85.6 & 192.2 & 89.3 & 73.4 \\ 
+ 3DPOF~\cite{xiang2019monocular} & 97.6 & 60.8 & 90.6 & 125.1 & 69.6 & 84.0 & 175.0 & 78.8 & 73.0 \\ 
+ EFT~\cite{joo2021exemplar} & 92.8 & 55.9 & 93.0 & 114.1 & 64.6 & 88.7 & 158.2 & 75.2 & 78.5 \\ 
+ 3DNBF & \textbf{88.8} & \textbf{53.3} & \textbf{93.6} & \textbf{109.4} & \textbf{62.2} & \textbf{90.9} & \textbf{150.2} & \textbf{72.0} & \textbf{85.3} \\ \bottomrule
\end{tabular}

\caption{\label{tab:optimization} \textbf{Comparison to optimization-base methods.} HMR-EFT is used for initialization. }
\end{table*}

\subsection{Training Setup and Datasets}

\noindent\textbf{Training.} Follow the common setting, we train NBV on \texttt{Human3.6M}~\cite{ionescu2013human3}, \texttt{MPI-INF-3DHP}~\cite{mehta2017monocular}, and \texttt{COCO}~\cite{lin2014microsoft} datasets. We use ground truth SMPL fittings for \texttt{Human3.6M} and \texttt{MPI-INF-3DHP}~\cite{kanazawa_hmr,kolotouros2019learning} and the pseudo-ground truth fittings from EFT~\cite{joo2021exemplar} for \texttt{COCO} following \cite{kocabas2021pare}. The selection of subjects for training strictly follows previous work~\cite{kolotouros2019learning,kocabas2021pare,joo2021exemplar}.
We first train the feature extractor on \texttt{COCO} for $175$K iterations, then fine-tune on all data for another $175$K iterations. During fine-tuning, the sampling ratio in each batch is $50\%$ \texttt{Human3.6M}, $20\%$ \texttt{MPI-INF-3DHP} and $30\%$ \texttt{COCO}. Note that for all baseline methods, we use the official model trained with the \textit{same} data as ours for fairness. \\

\noindent\textbf{Occlusion Robustness Evaluation. } We conduct evaluations on two datasets to measure the robustness and generalization of our method: an in-the-wild dataset \texttt{3DPW-Occ}~\cite{zhangoohcvpr20} which is a subset of the original \texttt{3DPW}~\cite{von2018recovering} dataset and an artificial indoor occlusion dataset \texttt{3DOH50K}~\cite{zhangoohcvpr20}. In particular, we directly test all models on these datasets \textit{without any training on them}. For \texttt{3DPW} and \texttt{3DPW-Occ}, we sample the videos every 30 frames. We report mean per joint position error (MPJPE) and Procrustes-aligned mean per joint position error (PA-MPJPE) in mm as the main evaluation metrics. We also report the 2D Percentage of Correct Keypoints with head length threshold (PCKh) to measure how well the prediction aligns with the 2D image. \\

\noindent\textbf{Adversarial occlusion robustness evaluation.} Inspired by the occlusion analysis in \cite{kocabas2021pare}, we design an adversarial protocol \texttt{3DPW-AdvOcc} to further evaluate the occlusion robustness of SOTA methods. Specifically, we slide an occlusion patch over the input image to find the worst prediction. This is done by comparing the relative performance degradation on the visible joints. We argue that evaluating the performance on occluded joints is sometimes ambiguous since the location of occluded joints is not always predictable even for human. Therefore, for a more stable and meaningful evaluation, the joints outside the bounding box or occluded by the patch are \textit{excluded from evaluation}. Instead of using a gray occlusion patch, we use textured patches generated by randomly cropping texture maps from the Describable Textures Dataset (DTD)~\cite{cimpoi2014describing}, which is more challenging. Two different square patch sizes are used: $40$ and $80$ relative to a $224\times 224$ image, denoted as Occ@40 and Occ@80 respectively, and the stride is set to $10$.

\subsection{Performance Evaluation}\label{sec:exp-performance}

\noindent\textbf{Baselines.} To demonstrate the superior performance and occlusion robustness of 3DNBF, we compare our model with four SOTA regression-based methods: SPIN~\cite{kolotouros2019learning}, HMR-EFT~\cite{joo2021exemplar}, Mesh Graphormer~\cite{lin2021mesh} and PARE~\cite{kocabas2021pare} where PARE is designed to be robust to occlusions with part attention and trained with synthetic occlusion augmentation. For fair comparisons, we adopt the models with the \textit{same} ResNet-50 backbone for all methods. We also compare 3DNBF with SOTA optimization-base methods that also improve occlusion robustness: SMPLify~\cite{bogo2016keep}, 3D POF~\cite{xiang2019monocular} and EFT~\cite{joo2021exemplar}. 

\noindent\textbf{Comparison to SOTA.} As shown in Table~\ref{tab:regressor}, we first evaluate on the standard \texttt{3DPW} test set and 3DNBF achieves SOTA performance. On occlusion datasets \texttt{3DPW-Occ} and \texttt{3DOH50K}, our improvement becomes more significant. We then evaluate the occlusion robustness on \texttt{3DPW-AdvOcc} where we find all regression-based methods suffer from occlusion with the MPJPE increasing up to $225\%$ and the PA-MPJPE increasing up to $127\%$ even for the best-performing method. The transformer-based model \cite{lin2021mesh} suffers the most from occlusion which we speculate to be due to overfitting. In contrast, 3DNBF is much more robust to occlusion improving over the SOTAs by a wide margin. Note that our predictions align better with the image as shown in PCKh.

\noindent\textbf{Comparison to other optimization-based methods.} We compare 3DNBF with three optimization-based methods on \texttt{3DPW} and \texttt{3DPW-AdvOcc}. We choose HMR-EFT as initial regressor to use the official EFT implementation. All methods use the same 2D keypoints detected by OpenPose~\cite{cao2017realtime}. As shown in Table~\ref{tab:optimization}, we achieve the best performance on both non-occluded and occluded settings. Although SMPLify improves 2D PCKh, it does not quite improve on the 3D metrics. This is due to SMPLify only fitting SMPL parameters to 2D keypoints without capturing any 3D information from the image, thus suffering from the 2D-3D ambiguity. EFT fine-tunes the regression network using a 2D keypoint reprojection loss. It achieves better performance than SMPLify because the regression network itself can implicitly encode 3D information in the input image and serve as a conditional 3D pose prior. However, EFT does not improve on non-occluded cases.

\noindent\textbf{Qualitative results.} We qualitatively demonstrate the improved occlusion robustness of 3DNBF compared to the SOTA regression-based method PARE in Fig.~\ref{fig:regressor} and show more comparisons in the supplementary material. Notice that PARE makes unaligned predictions for visible joints. 

\begin{figure}[t]
    \centering
    \includegraphics[width=\linewidth]{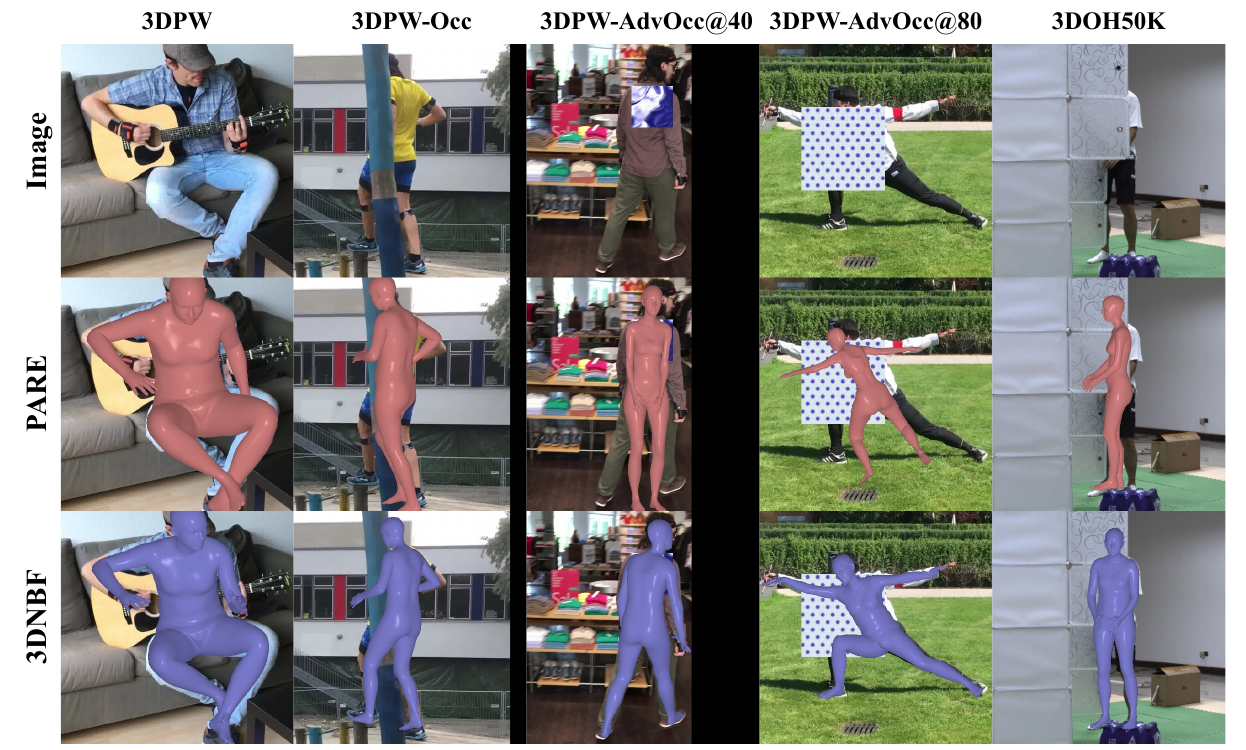}
    \caption{\textbf{Qualitative results on evaluated datasets. }
    \label{fig:regressor}}
    
\end{figure}

\subsection{Ablation Studies}
\label{subsec:ablation}
In this section, we provide ablation of different components in 3DNBF including the NBV, the design of pose-dependent features, and the 3D-aware contrastive loss. All experiments are on \texttt{3DPW-AdvOcc@80}. 

\begin{table}[]
\setlength\tabcolsep{2pt}
\begin{subtable}[t]{0.52\linewidth}
\scriptsize
\centering
\begin{tabular}[t]{@{}lccc}
\toprule
 & {\ssmall MPJPE$\downarrow$} & {\ssmall P-MPJPE$\downarrow$} & {\ssmall PCKh$\uparrow$} \\ \midrule
3DNBF & \textbf{140.7} & \textbf{71.8} & \textbf{85.0} \\ \midrule
Init. only  & 171.4 & 80.5 & 75.8 \\
w/o NBV  & 146.6 & 72.6 & 83.1 \\
w/o contrast & 167.4 & 80.4 & 79.4 \\
\bottomrule
\end{tabular}
\caption{\label{tab:ablation_component} Ablations for 3DNBF. }
\end{subtable}
\begin{subtable}[t]{0.45\linewidth}
\scriptsize
\centering
\begin{tabular}[t]{lccc}
\toprule
O & {\ssmall MPJPE$\downarrow$} & {\ssmall P-MPJPE$\downarrow$} & {\ssmall PCKh$\uparrow$} \\ \midrule
1 & 189.6 & 84.8 & 69.2 \\
4 & \textbf{140.7} & \textbf{71.8} & \textbf{85.0} \\
8 & 161.2 & 77.8 & 81.0 \\ \bottomrule
\end{tabular}
\caption{\label{tab:ablation_o}\# of pose-dependent features. $O\texttt{=}1$ means using pose-independent feature. }
\end{subtable}
\label{tab:ablation}
\caption{\textbf{Ablation studies.} All experiments are performed on \texttt{3DPW-AdvOcc@80}. (P-MPJPE: PA-MPJPE.)} 

\end{table}

\noindent\textbf{NBV vs. Mesh}. To demonstrate the advantage of NBV over mesh representation, we replace it with a mesh-based neural representation using SMPL while keeping everything else the same. We use the differentiable rendering implementation from PyTorch3D~\cite{ravi2020accelerating}. As shown in Table~\ref{tab:ablation_component}, this model achieves worse results than using NBV. 

\noindent\textbf{The pose-dependent kernel features. } The 3D-aware pose-dependent kernel feature is key to the success of 3DNBF. Here we validate its effectiveness by comparing it with pose-independent feature ($O\texttt{=}1$). As shown in Table~\ref{tab:ablation_o}, much better performance is achieved with $O\texttt{>}1$. Using $4$ features for each kernel achieves the best performance while further increasing the number of features may make the learning harder as it introduced more parameters. 

\noindent\textbf{Importance of contrastive training. } We ablate this by training our model without contrastive learning, i.e. training the feature extractor with regression loss only. The performance degrades a lot as shown in Table~\ref{tab:ablation_component}. The intuition is that our model requires the features to be Gaussian distributed and contrastive learning encourages this.

\noindent\textbf{Regression head performance. } Although we do not expect the regression head to be robust to occlusion, it achieves higher occlusion robustness compared to other regression-based methods as shown in Table~\ref{tab:ablation_component} and Table~\ref{tab:regressor}. 

\section{Conclusion}
In this work, we introduce 3D Neural Body Fitting (3DNBF) - an approximate analysis-by-synthesis approach to 3D HPE that is accurate and highly robust to occlusion. To this end, NBV is proposed which is an explicit volume-based generative model of pose-dependent features for human body. 
We propose a contrastive learning framework for training a feature extractor that captures the 3D pose information of the body parts thus overcoming the 2D-3D ambiguity in monocular 3D HPE. Experiments on challenging benchmark datasets demonstrate that 3DNBF outperforms SOTA regression-based methods as well as optimization-based methods. While focusing on occlusion robustness in this paper, we expect our model to be robust to other challenging adversarial examinations~\cite{shu2020identifying,ruiz2022simulated}.

\section{Acknowledgements}
AK acknowledges support via his Emmy Noether Research Group funded by the German Science Foundation (DFG) under Grant No. 468670075.
This research is based upon work supported in part by the Office of the Director of National Intelligence (ODNI), Intelligence Advanced Research Projects Activity (IARPA), via [2022-21102100005]. This work was also supported by NIH R01 EY029700, Army Research Laboratory award W911NF2320008, and Office of Naval Research N00014-21-1-2812.

{\small
\bibliographystyle{ieee_fullname}
\bibliography{egbib}
}
\end{document}